\documentclass[letterpaper]{article} 
\usepackage{aaai2026}  
\usepackage{times}  
\usepackage{helvet}  
\usepackage{courier}  
\usepackage[hyphens]{url}  
\usepackage{graphicx} 
\usepackage{natbib}  
\usepackage{caption} 
\usepackage{algorithm}
\usepackage{algpseudocode} 

\usepackage{booktabs}
\usepackage{multirow}
\usepackage{amsmath}
\usepackage{amssymb}

\usepackage[titletoc,title]{appendix}

\usepackage{newfloat}
\usepackage{listings}
\DeclareCaptionStyle{ruled}{labelfont=normalfont,labelsep=colon,strut=off} 
\floatstyle{ruled}
\newfloat{listing}{tb}{lst}{}
\floatname{listing}{Listing}
\title{DualFete: Revisiting Teacher-Student Interactions from a Feedback Perspective for Semi-supervised Medical Image Segmentation}
\author{
    Le Yi\textsuperscript{\rm 1}, 
    Wei Huang\textsuperscript{\rm 1}\footnotemark[1], 
    Lei Zhang\textsuperscript{\rm 1}\thanks{Corresponding author: Wei Huang; Lei Zhang. }, 
    Kefu Zhao\textsuperscript{\rm 1}, 
    Yan Wang\textsuperscript{\rm 2}, 
    Zizhou Wang\textsuperscript{\rm 2}
}
\affiliations{
    \textsuperscript{\rm 1}College of Computer Science, Sichuan University, China \\
    \textsuperscript{\rm 2}Institute of High Performance Computing, A*STAR, Singapore\\
    \{yile, kefuzhao\}@stu.scu.edu.cn, \{weihuang, leizhang\}@scu.edu.cn, \{wangyan, wang\_zizhou\}@a-star.edu.sg
}

\begin{document}

\maketitle

\begin{abstract}
The teacher-student paradigm has emerged as a canonical framework in semi-supervised learning. When applied to medical image segmentation, the paradigm faces challenges due to inherent image ambiguities, making it particularly vulnerable to erroneous supervision. Crucially, the student's iterative reconfirmation of these errors leads to self-reinforcing bias. While some studies attempt to mitigate this bias, they often rely on external modifications to the conventional teacher-student framework, overlooking its intrinsic potential for error correction. In response, this work introduces a feedback mechanism into the teacher-student framework to counteract error reconfirmations. Here, the student provides feedback on the changes induced by the teacher's pseudo-labels, enabling the teacher to refine these labels accordingly. We specify that this interaction hinges on two key components: the feedback attributor, which designates pseudo-labels triggering the student's update, and the feedback receiver, which determines where to apply this feedback. Building on this, a dual-teacher feedback model is further proposed, which allows more dynamics in the feedback loop and fosters more gains by resolving disagreements through cross-teacher supervision while avoiding consistent errors. Comprehensive evaluations on three medical image benchmarks demonstrate the method's effectiveness in addressing error propagation in semi-supervised medical image segmentation. 
\end{abstract}

\begin{links}
    \link{Code}{https://github.com/lyricsyee/dualfete}
\end{links}

\section{Introduction}
Medical image segmentation, which provides quantitative profiles for inner-body anatomical structures, plays a vital role in clinical practice and has emerged as a rapidly evolving subfield of AI for medicine~\cite{luo2022crossteach,chen2024transunet,li2025stitching,lan2025domain,lan2025loobox}. However, annotating medical images requires specialized expertise and is particularly labor-intensive at the voxel level. As a result, segmentation models often suffer from performance degradation due to the limited availability of labeled training data.

Semi-supervised learning (SSL) overcomes this limitation by leveraging additional supervision from unlabeled data. 
One line of SSL work follows the \textit{smoothness} assumption, which posits that if two samples are close in data space, then so should be their corresponding outputs~\cite{chapelle2009semi}. Consistency constraints, thus, are imposed on models to match the prediction of a perturbed sample to that of its vanilla counterpart~\cite{luo2021semi,huang2025gapmatch}. 
Another line follows the \textit{clustering} assumption, emphasising that samples should reside in high-density regions. It motivates pseudo-labeling unlabeled data for entropy minimization~\cite{grandvalet2004semi}; ultimately, samples can be compactly clustered by classes. 
Notably, accurate supervision is crucial for both lines of work, as models tend to perpetuate historical errors without awareness of their own mistakes, thereby rendering these errors increasingly difficult to correct. This degenerating case is known as \textit{confirmation bias}~\cite{ke2019dual}. 
\begin{figure}[tbp]
    \centering
    \includegraphics[scale=0.97]{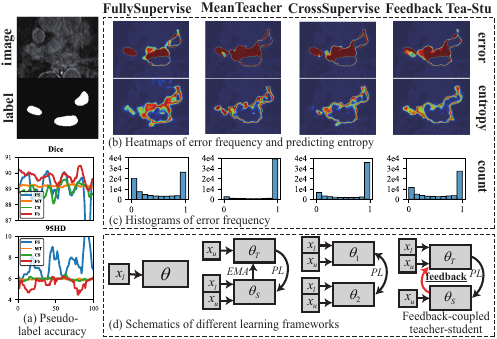}
    \caption{Pre-experiments on the LA dataset. 16 labeled samples are used.  FullySpervise runs with only the labeled set, while others use both. Results are evaluated on the first unlabeled sample and averaged from the last 100 training steps. It can be seen that medical image segmentation is susceptible to continual errors, so the confirmation bias issue is fairly problematic. Consistent errors can be reduced by the feedback interaction [\underline{highlighted} in (d)]. }
    \label{fig:schematic:study}
\end{figure}

This bias, unfortunately, incurs more hazards in semi-supervised medical image segmentation (SSMIS) due to image ambiguity; it leads to high regional uncertainty, especially near object boundaries. [Figure~\ref{fig:schematic:study}(b)]~\cite{yi2025learning}. For unseen samples, models are prone to generate erroneous pseudo-labels. Paradoxically, when trained on such error-prone supervision, they exhibit overconfidence in mistakes (low-entropy regions yet with high error frequency). Some studies attempt to prevent models from fitting to unreliable supervision, thereby partially alleviating error reinforcement~\cite{bai2023bidirectional,wu2023compete,wang2023mcf,shen2023co,chi2024adaptive}. However, these heuristic-based methods remain insufficient to counteract the intrinsic error-accumulation tendency of the prevailing teacher-student paradigm. This limitation stems from two fundamental issues. First, most methods rely on model-level perturbations to introduce teacher-student discrepancies and promote representation learning by resolving disagreement. However, such strategies are inadequate for highly nonlinear networks, as they tend to degenerate into self-training behavior, failing to sustain meaningful disagreement. Second, and more critically, the conventional teacher-student paradigm lacks inherent mechanisms for error correction. Although pseudo-labels can be influenced by the student using methods like Mean Teacher~\cite{tarvainen2017mean} or cross supervision~\cite{chen2021semi}[Figure~\ref{fig:schematic:study}(d)], the student  has no way to verify whether the update induced by pseudo-labels remains aligned with the constraint induced by labeled data. Once convergence, such methods yield vast consistent mistakes [Figure~\ref{fig:schematic:study}(c)] and also derive negligible change in pseudo-labeling accuracy [Figure~\ref{fig:schematic:study}(a)]. Thus, during the degenerative process toward self-training, they continually accumulate errors, exacerbating miscalibrated predictions and intensifying confirmation bias. 

In this work, we incorporate a feedback mechanism into the teacher-student framework, wherein the student assesses whether the updates driven by pseudo-labels align with the direction implied by ground-truth supervision. This assessment is then fed back to the teacher to guide refinements of pseudo-labels. Through this interaction, the framework acquires inherent error-correcting capabilities, helping to prevent error accumulation. As shown in Figure~\ref{fig:schematic:study}(c), this method effectively reduces recurring mistakes. To operationalize this feedback loop, we introduce two key components: the feedback attributor, which identifies the pseudo-labels responsible for triggering the student's update, and the feedback receiver, which determines the pseudo-labeling probabilities to be adjusted. Building on this foundation, we propose a dual-teacher feedback model, where two teachers collaboratively instruct a student and receive individualized feedback from the student, while also improving each other by resolving their mutual disagreements. This dual-teacher framework not only strengthens the feedback dynamics to mitigate persistent errors but also encourages constructive disagreement, fostering more effective teaching curricula. Extensive experiments are conducted on three benchmarks, including LA~\cite{xiong2021global}, Pancreas~\cite{roth2015deeporgan}, and BraTS~\cite{menze2014multimodal}, with sufficient justification of our method in these error-prone scenarios. 

Contributions are summarized as follows: 

(1) We introduce a feedback mechanism that equips the teacher-student model with error-correction capabilities, fundamentally mitigating vulnerability raised by error propagation in semi-supervised medical image segmentation.  

(2) We develop a dual-teacher feedback model that coordinates: (i) cross-supervision for resolving conflicts, and (ii) feedback loss for preserving teachers' diversity while suppressing errors. The two aspects enhance feedback dynamics and work collaboratively to teach a superior student. 

(3) Comprehensive experiments on three benchmarks confirm our method's effectiveness in combating this error-prone scenario, with sufficient justification through detailed quantitative and qualitative analysis. 

\section{Related Work}
\subsection{Semi-supervised medical image segmentation}
Advanced models are marching fastly and have made great breakthroughs in real-world segmentation tasks~\cite{dosovitskiy2020image,hu2023planning}. Such advancements, however, are lagging in the medical field mainly due to the scarcity of annotations. As a solution, researchers turn to SSL to excavate gains from limited labeled data and massive unlabeled data. Methods are categorized into two lines: consistency regularization~\cite{chapelle2009semi} and entropy minimization~\cite{grandvalet2004semi}. The former argues the output consistency between a perturbed data and its vanilla input. Such perturbations can be conducted at input~\cite{bai2023bidirectional,chi2024adaptive}, feature~\cite{yang2023revisiting,huang2024exploring}, and model~\cite{luo2022semi} levels. The latter argues for separating samples into clusters, driving to minimize entropy via pseudo labels. Representative methods, such as cross supervision~\cite{chen2021semi} and Co-training~\cite{qiao2018deep}, employ multiple learnable models to facilitate mutual learning. 

Notably, medical image ambiguity leads to error-prone supervision and inconsistent model predictions. 
While prior work has shown that models can benefit from learning courses designed to reduce ambiguity~\cite{xu2023ambiguity} and disagreement~\cite{shen2023co,wang2023mcf}, they inherently lack the ability to handle erroneous pseudo-labels in the teacher-student paradigm. Without explicit error-correction, this paradigm tends to accumulate more irreversible mistakes. 
Some studies address prediction disagreements while improving supervision accuracy, either by filtering unreliable targets using confidence thresholds~\cite{shen2023co,huang2025gapmatch} and predictive uncertainty~\cite{yu2019uncertainty,shi2021inconsistency}, or by competitively generating pseudo-labels from multiple models~\cite{wu2023compete,wang2023mcf,su2024mutual}. However, these solutions introduce additional architectural constraints that often fail to maintain meaningful disagreement due to the models' tendency to degenerate into trivial self-training behavior. 

This paper introduces a feedback mechanism within the teacher-student framework, endowing it with intrinsic error correction capabilities. With this foundation, we propose a dual-teacher feedback model that not only prevents error accumulation but actively benefits from disagreements. 

\subsection{Confirmation bias in semi-supervised learning}
In SSL, confirmation bias~\cite{nickerson1998confirmation} creates an error-reinforcing cycle where the model grows increasingly overconfident in its mistakes and resistant to correction~\cite{pseudoLabel2019}. 
\citet{rollwage2020confidence} claim that metacognitive interventions are one kind of method to combat the high confidence in the shaping process of confirmation bias in psychology. Such findings hint at least two routes to address this bias in SSL: (1) confidence reduction, which disrupts the reinforcing cycle of high-confidence errors, and (2) active interventions, which devise strategies to regularize the model toward expected learning direction. \citet{pseudoLabel2019} and \citet{liu2022acpl} supervise the model using soft pseudo-labels yielded by mixup techniques~\cite{zhang2018mixup}; they argue for the alleviation of high-confidence mistakes. Some studies impose a minimum labeled-data ratio, either per mini-batch~\cite{pseudoLabel2019} or per image~\cite{bai2023bidirectional}, to avoid the distributional shift far from the labeled data. Likewise, \citet{shen2023co} and \citet{chi2024adaptive} substitute uncertain patches with more reliable ones, and thus avoid accumulating too many errors. Some studies also propose competitive methods to ensure the correctness of pseudo-labels~\cite{wu2023compete,wang2023mcf}. 

While the above heuristically-driven solutions mitigate confirmation bias to some extent, the feedback interaction proposed in this work can address this issue more straightforwardly, as it inherently acts like a metacognitive intervention. Moreover, in a dual-teacher framework, we can enhance this feedback to be more dynamic. 

\section{Methodology}
Let $\theta$ be a segmentation model; it inputs an image $x$ and predicts probabilities $f^\theta := f(x;\theta)$. This model is expected to generalize well with a labeled dataset $\mathcal{D}_l= \{ (x^l, y^l) \}^{N_l}$ and an unlabeled dataset $\mathcal{D}_u= \{ x^u \}^{N_u}$, where $x^l$, $x^u$ represents the labeled and unlabeled sample, respectively, $y^l$ is the label mask of $x^l$, each voxel being one of $C:=\{0,...,C-1\}$ classes. $N_{l}=\left| \mathcal{D}_{l} \right|$, $N_{u}=\left| \mathcal{D}_{u} \right|$ is the respective dataset size, and it satisfies $N_l \ll N_u$. 

The teacher-student model is a learning paradigm in the sense that a teacher $\theta_{T}$ pseudo-labels unlabeled data $x^u$, which is used to supervise the student $\theta_{S}$ such that segmentation loss on $\mathcal{D}_u$ is minimized in terms of pseudo-labels $\hat{y}^u$: 
\begin{align}
    \hat{y}^u & = \mathop{\arg\max}_{c \in C} f_{c}(x^u;\theta_{T}) \label{eq:pseudo:label} \\
    \min_{\theta_{S}} \mathcal{L}_{S}(\theta_{S},  &\theta_{T}; \mathcal{D}_u) = \min_{\theta_{S}} \frac{1}{N_{u}} \sum_i^{N_u} \ell \left( f(x^{u}_{i};\theta_{S}), \hat{y}^u_{i} \right) \label{eq:student:object}
\end{align}
$f^{\theta}_{c}=f_c(\cdot;\theta)$ is the $c$-th class prediction, and $\ell$ is the loss function. Due to the ambiguity, medical image segmentation is susceptible to error supervision. 
However, the current teacher-student model can not perceive whether the update induced by pseudo-labels would further cause error reinforcement. 
To this end, we propose \textbf{DualFete}, which employs a feedback mechanism to equip the teacher-student model with error-correcting abilities. This mechanism will be further enhanced in a dual-teacher framework. 
\begin{figure}[tbp]
    \centering
    \includegraphics[scale=0.83]{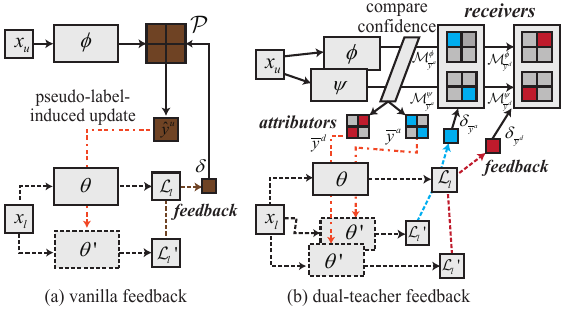}
    \caption{Schematic of the feedback mechanism. (a) Feedback is applied to each unit's likelihood, leading to a uniform updating direction. (b) The dual-teacher framework enables the feedback more dynamic based on prediction confidence. }
    \label{fig:schematic:feedback}
\end{figure}

\subsection{Feedback-coupled teacher-student model}
To avoid error accumulation, current SSMIS work sets a minimum labeled-data ratio per batch and employs some heuristically-driven methods to regularize training towards the accurate direction~\cite{yu2019uncertainty,luo2022semi}. Nonetheless, there is no way to verify if the updates induced by unlabeled data are aligned with those by labeled data. 

Motivated by this fact, performance change of the student attributed to the update via pseudo labels is quantified on labeled data. Intriguingly, if performance improved, pseudo labels yielded by the teacher are favorable, prompting further enhancement of pseudo-labeling; otherwise, teacher should reduce the likelihood of yielding such labels. This creates a feedback mechanism within the teacher-student model. In formal, given $\theta_{S}$, the performance on a labeled mini-batch $\mathcal{D}'_{l} \subset \mathcal{D}_{l}$  (with $N'_l = \left| \mathcal{D}'_{l} \right|$) is evaluated by
\begin{equation}
    \label{eq:loss:fs}
    \mathcal{L}_l(\theta_{S};\mathcal{D}'_{l}) = \frac{1}{N'_l} \sum^{N'_l}_{i} \ell \left( f\left( x^l_i; \theta_{S} \right), y^l_i  \right). 
\end{equation}
Let $\theta'_{S} = \theta_{S} - \eta \Delta \theta_{S}$ represents the student with one-step update, where $\Delta \theta_{S} = \nabla_{\theta_{S}} \mathcal{L}_{S}(\theta_{S},\theta_{T};\mathcal{D}'_u)$ is the stochastic gradient generated on an unlabeled mini-batch $\mathcal{D}'_u \subset \mathcal{D}_u$, and $\eta$ is the step size. Then, feedback can be derived: 
\begin{equation}
    \label{eq:vanilla:feedback}
    \delta = \mathcal{L}_l(\theta_{S}) - \mathcal{L}_l(\theta'_{S}). 
\end{equation} 
We drop $\mathcal{D}'_{l}$ from $\mathcal{L}_l$ for brevity. Let $\mathcal{P}\left(\hat{y}^u|x^u; \theta_{T}, \mathcal{D}'_u \right)$ represent the likelihood of the teacher generating pseudo-label as $\hat{y}^u$ on $\mathcal{D}'_u$. The teacher minimizes the following feedback loss $\mathcal{L}_{fb}$ to refine pseudo labels: 
\begin{equation}
    \mathcal{L}_{fb} (\theta_{T};\mathcal{D}'_u) = - \delta \log \mathcal{P}\left(\hat{y}^u|x^u; \theta_{T}, \mathcal{D}'_u \right). 
\end{equation}
Correspondingly, when $\delta > 0$, the teacher is guided to maximize the likelihood $\mathcal{P}$; while, when $\delta < 0$, the teacher is guided to reduce the likelihood. 

Theoretically, the feedback $\delta$ is a first-order approximation to the inner product of two gradients, i.e., $\Delta\theta_{S}$ and $\nabla_{\theta_{S}}\mathcal{L}_l(\theta'_{S})$, contained by a meta-objective~\cite{pham2021meta}. This means that when $\delta>0$, the update induced by pseudo-labels aligns with the gradient direction implied by the supervised loss $\mathcal{L}_{l}$; otherwise, the two gradients are in opposite directions. This meta-objective is in line with metacognitive intervention to address confirmation bias in psychology~\cite{rollwage2020confidence}. With this feedback mechanism, the teacher-student framework is equipped with inherent error-correcting abilities such that error accumulation can be circumvented.

\subsection{Dual-teacher feedback}
While the feedback loss $\mathcal{L}_{fb}$ provides signals for adjusting pseudo-labels $\hat{y}^u$, it enforces a uniform update direction across all voxel predictions, potentially limiting its corrective capacity [Figure~\ref{fig:schematic:feedback}(a)]. To address this, we propose an enhanced feedback mechanism within a dual-teacher framework, which not only enriches feedback dynamics but also promotes mutual learning between teachers [Figure~\ref{fig:schematic:feedback}(b)]. To this end, two key components are identified from $\mathcal{L}_{fb}$, namely the feedback attributor and receiver. Conceptually, the attributor refers to a set of pseudo-labels with which the student yields gradient for update, while the receiver indicates to which region feedback is applied to modulate the pseudo-labeling likelihood. In the following, the two components are specified in terms of the dual-teacher model. 

Let $\phi$, $\psi$ denote the two teachers, and $\hat{y}^{\phi_u}$, $\hat{y}^{\psi_u}$ be their predicted label by Eq.~\ref{eq:pseudo:label}. Pseudo-labels are set to the consensus if agreement is achieved, and set to the higher-confidence label if predictions conflict: 
\begin{equation}
\label{eq:dualt:pl}
\hat{y}^u = \begin{cases}
    \hat{y}^{\phi_u} \; \mbox{ or equivalently } \; \hat{y}^{\psi_u}, &  \mbox{if} \; \hat{y}^{\phi_u} = \hat{y}^{\psi_u}; \\ 
    \mathop{\arg\max}_c \max_{\phi, \psi} \{ f^\phi_c, f^\psi_c \},   &  \mbox{if} \; \hat{y}^{\phi_u} \ne \hat{y}^{\psi_u}. 
\end{cases}
\end{equation} 
This method is more likely to yield accurate pseudo-labels. 
Then, two types of feedback are quantified, induced by pseudo-labels of agreement and disagreement regions. We use $\bar{y}^{a} = \{ \hat{y}^u | \hat{y}^{\phi_u} = \hat{y}^{\psi_u} \}$, $\bar{y}^{d} = \{ \hat{y}^u | \hat{y}^{\phi_u} \ne \hat{y}^{\psi_u} \} $ to denote the attributor in terms of the agreement and disagreement region, respectively. Based on Eq.~\ref{eq:vanilla:feedback}, we have 
\begin{equation}
    \label{eq:dualt:fb}
    \delta_{\bar{y}} = \mathcal{L}_l \left( \theta_S \right) - \mathcal{L}_l  \left( \theta_{S} - \eta \frac{\Delta^{\bar{y}} \theta_{S}}{ \Vert \Delta^{\bar{y}} \theta_{S}  \Vert } \right).
\end{equation} 
Here, $\bar{y} \in \{ \bar{y}^{a}, \bar{y}^{d} \}$. $\Delta^{\bar{y}} \theta_{S}$ is the gradient of $\mathcal{L} _{S}$ w.r.t. $\theta_{S}$ induced by pseudo-labels $\bar{y}$. Without conflicts, we employ $\delta_{a}$, $\delta_{d}$, respectively, to simplify the notations $\delta_{\bar{y}^{a}}$, $\delta_{\bar{y}^{d}}$. 

In our design, the agreement feedback $\delta_{a}$ is applied to the lower-confidence side between $\phi$ and $\psi$. This is because when $\delta_{a} > 0$, we would like to maximize the confidence lower-bound of $\hat{y}^u$; when $\delta_{a} < 0$, the lower confidence favors the easier occurrence of disagreement between two teachers. On the contrary, the disagreement feedback $\delta_{d}$ is applied to the higher-confidence side. When $\delta_{d} > 0$, the pseudo-label $\hat{y}^u$ should become more confident, while when $\delta_{d} < 0$, $\hat{y}^u$ is more likely to flip to the label predicted by the other teacher. With this idea, feedback receivers can be identified. 
Given a teacher $\theta \in \{ \phi, \psi \}$, and let $\bar{\theta}$ be the other, the receiver mask is defined at first by $\mathcal{M}^{\theta}_{\bar{y}} := \mathcal{I} \left[ \hat{y}^{\theta_{u}} \star \hat{y}^{\bar{\theta}_u}, f_{\hat{y}^{\theta_u}}^{\theta} \diamond f_{\hat{y}^{\bar{\theta}_u}}^{\bar{\theta}} \right]$. $\mathcal{I}\left[ \cdot \right]$ is the indicator function. $\star$, $\diamond$ are two comparative operators depended on $\bar{y}$; they are respectively replaced by $=$, $<$ for $\bar{y}^{a}$, while by $\ne$, $>$ for $\bar{y}^{d}$. 
We can then define the receiver in terms of the feedback $\delta_{\bar{y}}$ as $\mathcal{P}\left(\hat{y}^{\theta_{u}}|x^u; \theta, \mathcal{D}_{u}, \mathcal{M}^{\theta}_{\bar{y}} \right)$, i.e., the pseudo-labeling likelihood of $\theta$ on $\mathcal{D}_{u}$ after masking regions by $\mathcal{M}^{\theta}_{\bar{y}}$. 
As a result, the dual-teacher feedback loss is formulated as
\begin{equation}
    \label{eq:dualfete:lossfb}
    \mathcal{L}_{df} (\theta) =  - \sum_{\bar{y} \in \{ \bar{y}^{a}, \bar{y}^{d} \}} \delta_{\bar{y}} \log \mathcal{P}\left(\hat{y}^{\theta_{u}}|x^u; \theta, \mathcal{D}_{u}, \mathcal{M}^{\theta}_{\bar{y}} \right). 
\end{equation}

This design gives rise to two distinct yet coordinated types of feedback, each exhibiting greater dynamics by adapting to changes in prediction confidence -- thereby breaking the limits of uniform updates. Meanwhile, the individualized feedback not only helps prevent teachers from making consistent errors, but also fosters productive disagreement between them, enabling mutual improvement through cross-supervision [Eq.~\ref{eq:loss:w2s:cs}]~\cite{qiao2018deep,shen2023co}. 

\subsection{Holistic framework}
In total, the training procedure for a stochastic step is presented in Algorithm~\ref{alg:dualfete}. 
There exist differences in the training data and objectives between the teacher and the student. 

\begin{algorithm}[tbp]
    \caption{DualFete in a stochastic training step}
    \label{alg:dualfete}
    \renewcommand{\algorithmicrequire}{\textbf{Input:}}
    \renewcommand{\algorithmicensure}{\textbf{Models:}}
    \begin{algorithmic}[1]
        \Require Randomly sampled mini-batches of labeled data $\mathcal{D}'_{l} \subset \mathcal{D}_{l}$ and unlabeled data $\mathcal{D}'_{u} \subset \mathcal{D}_{u}$. 
        \Ensure Two teachers $\phi$, $\psi$, and a student $\theta_S$
        \State Pseudo-label $\mathcal{D}'_{u}$ by $\phi$, $\psi$, respectively.  \Comment [Eq.~\ref{eq:pseudo:label}]
        \State Get pseudo labels $\hat{y}^u$ and attributors $\bar{y}^a$, $\bar{y}^d$. \Comment [Eq.~\ref{eq:dualt:pl}]
        \State Evaluate feedback $\delta_{\bar{y}^a}$, $\delta_{\bar{y}^d}$. \Comment [Eq.~\ref{eq:dualt:fb}]
        \State Get receiver masks $\mathcal{M}^{\phi}_{\bar{y}^a}$, $\mathcal{M}^{\phi}_{\bar{y}^d}$, $\mathcal{M}^{\psi}_{\bar{y}^a}$, and $\mathcal{M}^{\psi}_{\bar{y}^d}$. 
        \State Compute dual-teacher feedback loss $\mathcal{L}_{df}$. \Comment [Eq.~\ref{eq:dualfete:lossfb}]
        \State Update $\theta_S$ by $\mathcal{L}_{S}$ using $\mathcal{D}'_{u}$ and $\hat{y}^u$. \Comment [Eq.~\ref{eq:student:object}]
        \State Update $\{\phi, \psi\}$ by $\mathcal{L}_T(\phi)+\mathcal{L}_T(\psi)$ using both $\mathcal{D}'_{l}$ and $\mathcal{D}'_{u}$. \Comment [$\mathcal{L}_{df}$ and Eq.~\ref{eq:teacher:loss},~\ref{eq:loss:w2s:cs}]
    \end{algorithmic}
    \Return $\phi$, $\psi$, and $\theta_S$
\end{algorithm}

\noindent
\textbf{Student Model.} The student updates its parameters only using $\mathcal{D}_u$ with pseudo-label $\hat{y}^u$ yielded by the dual-teacher model [Eq.~\ref{eq:dualt:pl}, Eq.~\ref{eq:student:object}]. Except for learning on $\mathcal{D}_u$, the student takes the duty of feeding back the performance change on labeled data $\mathcal{D}_l$ per-step of training [Eq.~\ref{eq:dualt:fb}]. (Optionally) The student can be further fine-tuned with $\mathcal{D}_l$, particularly when the gap between $N_l$ and $N_u$ is relatively small. 

\begin{table*}
    \centering
    \begin{tabular}{c|ccc|cc|cc}
    \hline\hline
    \multirow{1.5}[2]{*}{Methods} & \multicolumn{3}{c|}{\textbf{LA}} & \multicolumn{2}{c|}{\textbf{Pancreas}} & \multicolumn{2}{c}{\textbf{BraTS}} \\
        & 5\% (4) & 10\% (8) & 20\% (16) & 10\% (6) & 20\% (12) & 10\% (25) & 20\% (50) \\
    \hline
    \multicolumn{1}{l|}{Fully Supervise} & 52.55, 47.1 & 82.74, 13.4 & 86.96, 11.9 & 55.60, 45.3 & 72.38, 19.4 & 74.43, 37.1 & 80.16, 22.7 \\
    \hline
    \multicolumn{1}{l|}{UA-MT [Y.~\citeyearpar{yu2019uncertainty}]} & 82.26, 13.7 & 86.28, 18.7 & 88.74, 8.39 & 66.44, 17.0 & 76.10, 10.8 & 84.64, 10.5 & 85.32, 8.68 \\
    \multicolumn{1}{l|}{SASSNet [L.~\citeyearpar{li2020shape}]} & 81.60, 16.2 & 85.22, 11.2 & 89.16, 8.95 & 68.97, 18.8 & 76.39, 11.1 & 84.73, 9.88 & 85.64, 9.17 \\
    \multicolumn{1}{l|}{DTC [L.~\citeyearpar{luo2021semi}]} & 81.25, 14.9 & 87.51, 8.23 & 89.52, 7.07 & 66.58, 15.5 & 76.27, 8.70 & -  & - \\
    \multicolumn{1}{l|}{URPC [L.~\citeyearpar{luo2022semi}]} & 82.48, 14.7 & 85.01, 15.4 & 88.74, 12.7 & 73.53, 22.6 & 80.02, 8.51 & 84.53, 9.79 & 85.38, 8.36 \\
    \multicolumn{1}{l|}{PS-MT [L.~\citeyearpar{liu2022perturbed}]} & 88.49, 8.12 & 89.72, 6.94 & 90.02, 6.74 & 76.94, 13.1 & 80.74, 7.41 & 84.88, 9.93 & 85.91, 8.63 \\
    \multicolumn{1}{l|}{MC-Net+ [W.~\citeyearpar{wu2022mutual}]} & 83.59, 14.1 & 88.96, 7.93 & 91.07, 5.84 & 70.00, 16.0 & 80.59, 6.47 & 84.96, 9.45 & 86.02, 8.74 \\
    \multicolumn{1}{l|}{SS-Net [W.~\citeyearpar{wu2022exploring}] } &  86.33, 9.97 & 88.55, 7.49 & 89.28, 7.29 & 71.76, 17.6 & 78.98, 8.86 & - & - \\
    \multicolumn{1}{l|}{BCP [B.~\citeyearpar{bai2023bidirectional}]} & 88.02, 7.90 & 89.62, 6.81 & 91.26, 5.76 & 73.83, 12.7 & 82.91, 6.43 & 85.14, 9.89 & 86.13, 8.99 \\
    \multicolumn{1}{l|}{UniMatch [Y.~\citeyearpar{yang2023revisiting}]} & - & 89.04, 7.26 & 90.99, 6.07 & - & 82.35, 7.66 & 85.03, 9.50 & 85.84, 8.68 \\
    \multicolumn{1}{l|}{MutRel [S.~\citeyearpar{su2024mutual}]} & 87.20, 9.90 & 89.86, 6.91 & 91.02, 5.78 & 75.93, 9.07 & 81.53, 6.81 & 84.29, 9.57 & 85.47, \textbf{7.76} \\ 
    \multicolumn{1}{l|}{AD-MT [Z.~\citeyearpar{zhao2024alternate}]} & 89.63, 6.56 & 90.55, 5.81 & -     & 80.21, 7.18 & 82.61, 4.94 & -     & - \\ 
    \multicolumn{1}{l|}{TraCoCo [L.~\citeyearpar{liu2024translation}]} & - & 89.86, 6.81 & 91.51, 5.63 & 79.22, 8.46 & 83.36, 7.34 & 85.71, 9.20 & \textbf{86.69}, 8.04 \\
    \hline
    DualFete [ours] & \textbf{90.35}, \underline{6.42} & \textbf{91.28}, \underline{5.51} & \underline{91.89}, \underline{5.24} & \underline{81.99}, \textbf{5.34} & \underline{83.49}, \underline{4.76} & \underline{86.13}, \underline{9.02} & 85.83, 8.12 \\
    DualFete w.ft. [ours] & \underline{90.22}, \textbf{5.89} & \underline{91.12}, \textbf{5.44} & \textbf{91.91}, \textbf{5.22} & \textbf{82.45}, \underline{5.96} & \textbf{83.85}, \textbf{4.43} & \textbf{86.25}, \textbf{8.94} & \underline{86.46}, \underline{7.80} \\
    \hline\hline
\end{tabular}%
\caption{Comparison with SOTAs on LA, Pancreas, and BraTS19 datasets. The \textbf{best} and \underline{second best} results are highlighted. }
\label{tab:sotas}
\end{table*}

\noindent
\textbf{Dual-teacher Model.} Feedback from the student alone cannot guarantee superior teachers; they need to improve themselves through both individual and mutual learning. Thus, in addition to $\mathcal{L}_{df}$, the teachers are fully-supervised on $\mathcal{D}_l$ and cross-supervised on $\mathcal{D}_u$. Specifically, given a teacher $\theta \in \{ \phi, \psi \}$, the objective is formulated as 
\begin{equation}
    \label{eq:teacher:loss}
    \min_{\theta} \mathcal{L}_{T}(\theta) = \min_{\theta} \mathcal{L}_{l}(\theta) + \mathcal{L}_{df}(\theta) + \lambda \mathcal{L}^{\mathcal{A}}_{cs}(\theta; \bar{\theta}, \mathcal{A}). 
\end{equation}
Here, $\lambda$ is a ramp-up weighted factor, and $\mathcal{A}$ is a strong augmentator [e.g., copy-paste~\cite{ghiasi2021simple}; color-jittor~\cite{cubuk2020randaugment}]. $\mathcal{L}_{l}$ is the fully-supervised loss [Eq.~\ref{eq:loss:fs}], and the cross-supervised loss $\mathcal{L}^{\mathcal{A}}_{cs}$ is defined by 
\begin{equation}
    \label{eq:loss:w2s:cs}
    \mathcal{L}^{\mathcal{A}}_{cs}(\theta; \bar{\theta}, \mathcal{A}) = \frac{1}{N_{u}} \sum_i^{N_u} \ell \left( f\left(\mathcal{A}\left( x^{u}_{i} \right);\theta \right), \mathcal{A}\left(\hat{y}^{\bar{\theta}_{u}}\right) \right)
\end{equation}
Here, $\hat{y}^{\bar{\theta}_{u}}$ is the pseudo-label yielded by the teacher $\bar{\theta}$, and $\mathcal{A}(y)$ only operates positional transformations for the label $y$. In line with prior work, a confidence threshold is employed to filter possibly unreliable targets~\cite{yang2023revisiting,huang2025gapmatch}. Besides, we use $\mathcal{L}_{cs}$ to refer cross-supervised loss without weak-to-strong consistency. 

\begin{figure}[tbp]
    \centering
    \includegraphics[scale=0.75]{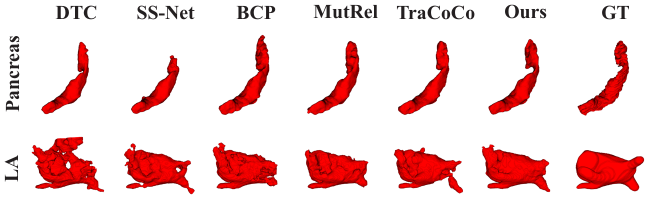}
    \caption{Visualizations of several methods (10\% labels). }
    \label{fig:vis:pan}
\end{figure}
\section{Experiments}
\subsection{Experimental settings}
\noindent 
\textbf{Datasets.} 
Experiments are conducted on LA~\cite{xiong2021global}, Pancreas~\cite{roth2015deeporgan}, and BraTS19~\cite{menze2014multimodal} datasets. We follow preprocessing steps and data split used in prior work~\cite{yu2019uncertainty,li2020shape,luo2021semi,liu2024translation,zhao2024alternate}. \textbf{LA} contains 100 MRIs, with 80 for training and 20 for testing. Three commonly used label-settings are involved, i.e., 5\%, 10\%, and 20\%. \textbf{Pancreas} contains 82 CT scans, in which 62 for training and 20 for testing. Two label settings are used 10\%, and 20\%. \textbf{BraTS} contains 335 brain MRIs, with 250 for training, 25 for validation, and 60 for testing. Label settings are the same as the Pancreas dataset. 

\noindent
\textbf{DualFete details.} We employ V-Net for LA and Pancreas, U-Net for BraTS, with the input size and evaluation protocol remaining the same as prior work~\cite{liu2024translation}. The SGD optimizer and data loading configurations are also in line with prior work~\cite{wu2022exploring,liu2024translation}. We employ the Dice and cross-entropy combination as the segmentation loss while using cross-entropy loss to evaluate student performance on labeled data. The likelihood $\mathcal{P}$ is implemented as the cumulative product of per-voxel pseudo-labeling probability. For LA, we employ normalized gradients in Eq.~\ref{eq:dualt:fb} while Pancreas and BraTS do not, as we experimentally find it works better. Moreover, we find that using likelihood of strong-augmented data can also be beneficial in some cases. Performance is evaluated by Dice (\%) and 95\% Hausdorff distance (95HD, voxel), displayed on the left and right sides of each table cell, respectively. 

\begin{figure*}[tbp]
    \centering
    \includegraphics[scale=0.9]{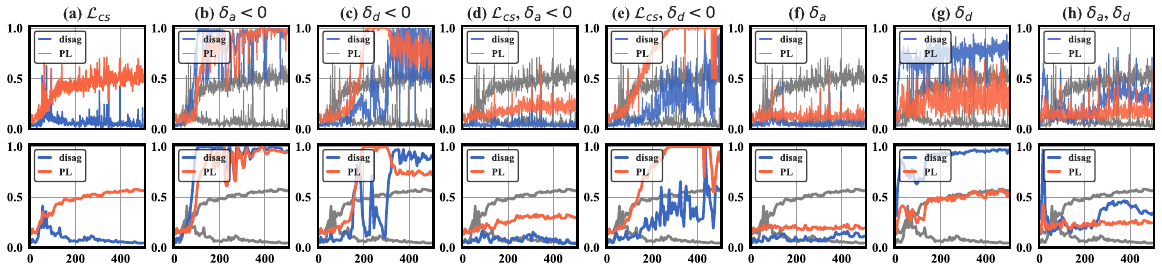}
    \caption{Disagreement between two teachers (disag) and the pseudo-label error (PL) [Eq.~\ref{eq:dualt:pl}], measured by 1-Dice. We report results evaluated by training inputs (first row) and by the testing set (last row), respectively. (LA, 10\% labels). }
    \label{fig:abl:fb:quali}
\end{figure*}
\subsection{Comparison with SOTAs}
We first compare DualFete with the fully supervision baseline (only with $\mathcal{D}_{l}$) and various state-of-the-art methods, including: UA-MT~\cite{yu2019uncertainty}, SASSNet~\cite{li2020shape}, DTC~\cite{luo2021semi}, URPC~\cite{luo2022semi}, PS-MT~\cite{liu2022perturbed}, MC-Net+~\cite{wu2022mutual}, SS-Net~\cite{wu2022exploring}, BCP~\cite{bai2023bidirectional}, UniMatch~\cite{yang2023revisiting}, MutRel~\cite{su2024mutual}, AD-MT~\cite{zhao2024alternate}, and TraCoCo~\cite{liu2024translation}. The reported results of these methods are identical to those in \citet{su2024mutual}, \citet{zhao2024alternate}, and \citet{liu2024translation} under the same settings. Our results are from the student, and we also report the fine-tuned performance (DualFete w.ft.). 

While recent state-of-the-art methods [AD-MT~\cite{zhao2024alternate}, TraCoCo~\cite{liu2024translation}] have made significant progress in SSMIS, our DualFete pushes these boundaries further. As shown in Table~\ref{tab:sotas}, DualFete outperforms existing approaches across nearly all benchmarks and label settings, even without fine-tuning. Notably, on Pancreas with 10\% labels, it achieves a +1.78\% Dice improvement and 1.84-voxel 95HD reduction over previous best results [AD-MT~\cite{zhao2024alternate}]. This consistent outperformance suggests that our framework captures fundamental improvements beyond current teacher-student paradigms. The purely unlabeled training of the student model makes it possible to yield additional gains through fine-tuning on labeled data. These gains are substantial in the 20\% labeled data setting and on the more challenging Pancreas dataset. However, fine-tuning on very limited data is susceptible to overfitting, e.g., on LA with 5\% and 10\% labels. Thus, DualFete has particular potential in label-scarce cases by using labeled data exclusively to guide pseudo-label updates toward correct directions, thereby avoiding the need for fine-tuning and circumventing the overfitting issue. For the BraTS dataset, it is noteworthy that our DualFete shows some oscillations in terms of testing performance. We speculate that the validation set is prone to overfitting due to its relatively limited quantity (25 samples).  

Figure~\ref{fig:vis:pan} illustrates two challenging segmentation cases in the Pancreas and LA dataset. Our method shows the best alignments with the ground-truth masks, maintaining most spatial details, which suggests the improved generalizability. 

\subsection{Ablation studies}
\textbf{Qualitative analysis of DualFete.} To investigate the effectiveness of DualFete, we pretrain a cross-supervised model ($\mathcal{L}_{l}+\mathcal{L}_{cs}$) for 6k steps and then experimentally impose different constraints with 0.5k extra steps. We evaluate the pseudo-label error and the disagreement between two teachers in the second stage. 

Several findings can be summarized by Figure~\ref{fig:abl:fb:quali}. 
\textbf{(1)} The cross-supervised loss $\mathcal{L}_{cs}$ forces consensus between the teachers but amplifies errors, leading to consistently worse pseudo-labels [Figure~\ref{fig:abl:fb:quali}(a)]. \textbf{(2)} The constraint of $\delta_{a} < 0$ forces teachers to fully conflict with each other [Figure~\ref{fig:abl:fb:quali}(b)], while $\mathcal{L}_{cs}$ enables them to resolve these conflicts while reducing pseudo-labeling errors [Figure~\ref{fig:abl:fb:quali}(d)]. \textbf{(3)} The $\delta_{d} < 0$ constraint induces pseudo-labeling oscillation between the teachers. In extreme cases, this creates collapsing dynamics where object boundaries predicted by the teachers are alternatively eroded, resulting in background-only predictions ultimately. This collapse can be observed from the completely incorrect pseudo-labels yet relatively low disagreement [Figure~\ref{fig:abl:fb:quali}(c,e)]. 
Nonetheless, the observation suggests that if the constraint is prevented from collapsing, it has the potential to avoid invariant pseudo-labels. \textbf{(4)} $\delta_{d} > 0$ can foster disagreements by reinforcing pseudo-label confidence [Figure~\ref{fig:abl:fb:quali}(g)]. Since the student is trained from scratch, it generates predominantly positive feedback, which continually reinforces one teacher's confidence while relatively weakening the other. Ultimately, two teachers evolve complementary prediction behaviors. \textbf{(5)} The two types of feedback $\delta_{a}$, $\delta_{d}$ collaboratively counteract degenerating cases found in (2)-(4) [Figure~\ref{fig:abl:fb:quali}(h)]. 

These findings suggest that the two types of feedback operate in different manners, yet work synergistically in the dual-teacher model: they (1) generate productive prediction disagreements; (2) maintain pseudo-label accuracy while preventing error accumulation. 

\begin{table}
    \centering
    \begin{tabular}{cccc|c|c}
    \hline\hline
    D.T.  & Att. &  Rec. & $\mathcal{A}$ & LA 20\% & Pancreas 20\% \\
    \hline
    $\times$ & $\times$   & $\times$  & $\times$ & 88.55, 8.58 & 77.18, 9.81 \\
    $\times$ & $\hat{y}^u$ & $\mathcal{P}$ & $\times$ & 89.63, 7.92 & 79.27, 9.90 \\
    $\surd$  & $\hat{y}^u$ & $\mathcal{P}$ & $\times$ & 89.83, 8.08 & 76.83, 12.2 \\
    $\surd$  & $\bar{y}^{a}$ & $\mathcal{P}_{l}$ & $\times$ & 90.34, 6.43 &  79.56, 8.38 \\
    $\surd$  & $\bar{y}^{d}$ & $\mathcal{P}_{h}$ & $\times$ & \underline{90.35}, \underline{6.19} &  \underline{80.77}, 8.22 \\
    $\surd$  & $\bar{y}^{a,d}$ & $\mathcal{P}_{h,l}$ & $\times$ & 87.69, 8.73 & 78.06, 8.77 \\
    $\surd$  & $\bar{y}^{a,d}$ & $\mathcal{P}_{l,h}$ & $\times$ & \textbf{90.89}, \textbf{6.11} & \textbf{81.12}, \textbf{7.74} \\
    \hline
    $\times$ & $\hat{y}^u$ & $\mathcal{P}$ & $\surd$  & 89.12, 8.68 & 78.59, 8.68 \\
    $\surd$  & $\bar{y}^{a,d}$ & $\mathcal{P}_{l,h}$ & $\surd$  & 90.14, 7.30 & \underline{80.77}, \underline{7.89} \\
    \hline\hline
    \end{tabular}%
    \caption{Quantitative analysis of DualFete. $\hat{y}^u$ is either from Eq.~\ref{eq:pseudo:label} or Eq.~\ref{eq:dualt:pl} depended on whether the dual-teacher model is introduced (first column). In the second and third columns, $\bar{y}^{a,d}$ and $\mathcal{P}_{l,h}$ indicate that the update is triggerred by \underline{\textbf{a}}greement/\underline{\textbf{d}}isagreement pseudo-labels, and feedback is applied correspondingly to the \underline{\textbf{l}}ower/\underline{\textbf{h}}igher confidence side. $\mathcal{A}$ indicates whether $\mathcal{P}$ is from strong-augmented data. }
    \label{tab:fb:ablat}
\end{table}

\noindent
\textbf{Quantitative analysis of DualFete.} 
We further explore the proposed feedback mechanism quantitatively, and results are shown in Table~\ref{tab:fb:ablat}. Compared to the baseline teacher-student model (first row), the feedback mechanism actively prompts the teacher to refine pseudo-labels, yielding consistent student performance gains (2, 4, 5, and 7-th rows). The single-teacher framework relies solely on naive feedback, while the dual-teacher extension introduces two distinct and collaborative feedback types ($\delta_{a}$ and $\delta_{d}$), each capable of operating independently while surpassing the baseline. This extension achieves significant improvements over the single-teacher protocol, demonstrating the value of coordinated yet independent feedback dynamics. However, performance degrades markedly when feedback attributors and receivers are mismatched (3 and 6-th rows), highlighting the criticality of proper component alignment. Strong-augmented likelihoods here do not result in performance gains, as only when weak-to-strong consistency is imposed can teacher models learn to handle such cases, which will be shown later. 

\begin{table}
    \centering
    \begin{tabular}{c|c|c|c}
    \hline\hline
    loss  & teacher & Ep.1 (Dice) & Ep.2 (Entropy$\times10^4$) \\
    \hline
    \multirow{1.5}[2]{*}{+$\mathcal{L}^\mathcal{A}_{cs}$} 
    & $\phi$ & 78.42 $\pm$ 8.46 & 12.65 $\pm$ 2.45 \\
    & $\psi$ & 80.51 $\pm$ 6.23 & 13.13 $\pm$ 2.82 \\
    \hline
    \multirow{1.5}[2]{*}{+$\mathcal{L}_{df}$} 
        & $\phi$ & 57.40 $\pm$ 24.63 & 14.30 $\pm$ 2.77  \\
        & $\psi$ & 54.38 $\pm$ 28.83 & 15.23 $\pm$ 2.93  \\
    \hline\hline
    \end{tabular}%
    \caption{Experiments with either $\mathcal{L}_{df}$ only or $\mathcal{L}^\mathcal{A}_{cs}$ only. Label setting: Pancreas 20\%. }
    \label{tab:abl:cr:em}
\end{table}
\noindent
\textbf{Cross-teacher supervision.} 
We design two experiments to verify that the feedback loss $\mathcal{L}_{df}$ operates distinctly from both consistency regularization and entropy minimization. Specifically, we train two dual-teacher frameworks under identical configurations but with different unsupervised constraints: one using only feedback loss $\mathcal{L}_{df}$, and the other employing cross-supervised loss $\mathcal{L}^{\mathcal{A}}_{cs}$. In Experiment 1 (Ep.1), each test sample is evaluated six times with different strong augmentations, and we report the mean and standard deviation of Dice scores. Experiment 2 (Ep.2) similarly conducts six evaluations per test sample, yet with dropout perturbations. We measure the mean and standard deviation of per-image entropy summation. 
Table~\ref{tab:abl:cr:em} suggests two key insights of using $\mathcal{L}_{df}$ alone: (1) it fails to maintain robustness against input perturbations, evidenced by significant Dice degradation and high performance variance; (2) it increases prediction uncertainty, showing no reduction in output entropy. 

\begin{table}
    \centering
    \begin{tabular}{c|l|cc}
    \hline\hline
    dataset & \multicolumn{1}{c|}{loss}  & 10\%  & 20\% \\
    \hline
    \multirow{4.5}[2]{*}{LA} & $\mathcal{L}_{l}$ & 86.82, 9.96 & 88.55, 8.58 \\
        & +$\mathcal{L}_{cs}$ & 88.37, 9.27 & 90.30, 6.80 \\
        & +$\mathcal{L}_{cs}$+$\mathcal{L}_{df}$ & \underline{90.77}, \underline{6.19} & \underline{91.66}, \underline{6.12} \\
        & +$\mathcal{L}^{\mathcal{A}}_{cs}$ & 90.39, 7.03 & 91.10, \underline{6.12} \\
        & +$\mathcal{L}^{\mathcal{A}}_{cs}$+$\mathcal{L}_{df}$ & \textbf{91.28}, \textbf{5.51} & \textbf{91.89}, \textbf{5.24} \\
    \hline
    \multirow{4.5}[2]{*}{Pancreas} & $\mathcal{L}_{l}$ & 70.81, 14.3 & 77.18, 9.81 \\
        & +$\mathcal{L}_{cs}$ & 75.39, 10.6 & 79.62, 6.84 \\
        & +$\mathcal{L}_{cs}$+$\mathcal{L}_{df}$ & 78.13, 7.71 & 80.92, 7.89 \\
        & +$\mathcal{L}^{\mathcal{A}}_{cs}$ & \underline{80.30}, \underline{6.06}  & \underline{82.17}, \underline{5.43} \\
        & +$\mathcal{L}^{\mathcal{A}}_{cs}$+$\mathcal{L}_{df}$ & \textbf{81.99}, \textbf{5.34} & \textbf{83.49}, \textbf{4.76} \\
    \hline\hline
    \end{tabular}%
    \caption{Ablation study of $\mathcal{L}_{df}$, $\mathcal{L}_{cs}$, and $\mathcal{L}^\mathcal{A}_{cs}$ in the dual-teacher framework. Results are evaluated by the student. }
    \label{tab:abl:csup}
\end{table}
For further validation, we conducted an ablation study examining the impact of relevant unsupervised constraints. As shown in Table~\ref{tab:abl:csup}, cross-supervision between the dual teachers produces superior student performance, indicating that the teachers generate higher-quality pseudo-labels through mutual learning [$\mathcal{L}_{cs}$, $\mathcal{L}^{\mathcal{A}}_{cs}$]. This virtuous cycle is further amplified by student feedback [$\mathcal{L}_{df}$], which continuously improves teacher pseudo-labeling accuracy, thereby enabling progressive student performance gains. 

\begin{figure}[tbp]
    \centering
    \includegraphics[scale=0.53]{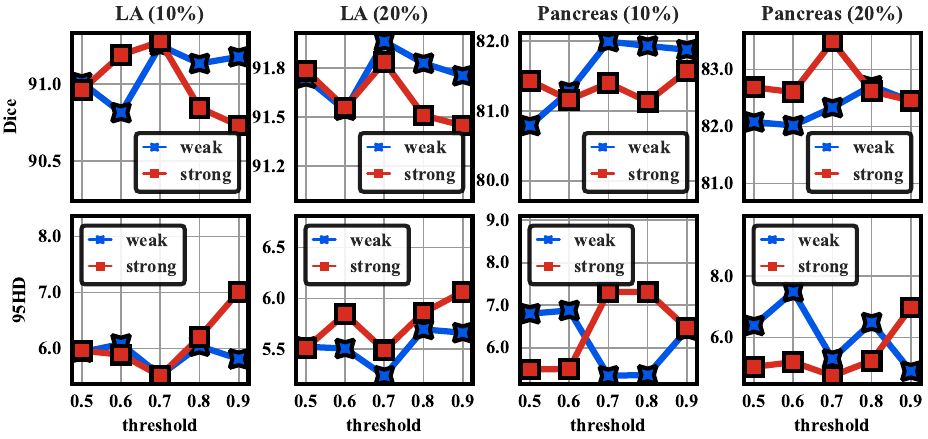}
    \caption{Effects of confidence thresholding (in $\mathcal{L}^{\mathcal{A}}_{cs}$) and strong-augmentation likelihood (in $\mathcal{L}_{df}$). }
    \label{fig:abl:thd}
\end{figure}
We experiment with varying confidence thresholds in cross-supervised loss $\mathcal{L}^{\mathcal{A}}_{cs}$ [Eq.~\ref{eq:loss:w2s:cs}] and analyze the effect of the likelihood type used in the feedback loss $\mathcal{L}_{df}$ [Eq.~\ref{eq:dualfete:lossfb}]. Figure~\ref{fig:abl:thd} demonstrates that a 0.7 threshold achieves optimal performance across most scenarios, validating the importance of filtering low-confidence predictions, which are possibly incorrect. More importantly, preventing converging to low-confidence targets further ensures achieving expected performance using feedback loss, as it mostly modulates the likelihood of these easy-to-error targets, thereby circumventing error accumulation. 
Notably, strong-augmentation likelihood yields significant improvements on the Pancreas 20\% benchmark and also remains effective with lower thresholds. We hypothesize this occurs because teachers benefit from additional low-confidence supervision to better align the likelihood from strong and weak augmentations. 

\begin{table}
    \centering
    \begin{tabular}{c|c|c|c}
    \hline\hline
    Method & Train (s/iter)  & Mem. (GB) & Infer. (s/case) \\
    \hline
    FullySup & 0.15  & 5.15  & 1.93 \\
    AD-MT & 0.67  & 7.26  & 1.93 \\
    TraCoCo & 2.39  & 21.93 & 1.81 \\
    DualFete & 2.28  & 10.25 & 1.91 \\
    \hline\hline
    \end{tabular}%
    \caption{Training (time, memory) and inference (time) cost on the LA dataset with V-Net as the backbone. }
    \label{tab:abl:eff}
\end{table}

\noindent
\textbf{Efficiency.} The dual-teacher design is used only during training, while inference relies solely on the student. We consider the trade-off worthwhile between training cost and inference performance. As show in Table~\ref{tab:abl:eff}, our method achieves comparable inference speed (about 1.9 s/case with TITAN RTX) with superior accuracy (see in Table~\ref{tab:sotas}). 

\section{Conclusions}
In this work, we revisit the teacher-student paradigm through the lens of feedback and propose a dual-teacher feedback model to address confirmation bias in semi-supervised medical image segmentation. Our framework makes two key contributions: (1) an error-correcting feedback mechanism that mitigates error propagation, and (2) a collaborative architecture that integrates the student's individualized feedback with mutual refinement between teachers, thereby enhancing teaching effectiveness. The dual-teacher design strengthens feedback dynamics, enabling both productive disagreement between teachers and reduced error consistency, two essential factors for improving robustness under imperfect supervision. Extensive experiments on three benchmark datasets demonstrate consistent performance gains, with both quantitative metrics and qualitative analyses validating the effectiveness of our approach.

\section{Acknowledgments}
This work was supported in part by the National Natural Science Foundation for Distinguished Young Scholar under Grant No.62025601, in part by the National Natural Science Foundation Regional Innovation and Development Joint Fund under Grant No.U24A20341, in part by Transformation of Tianfu Jincheng Laboratory under Grant No.2025ZH013, and in part by Sichuan Province Innovative Talent Funding Project for Postdoctoral Fellows under Grant No.BX202512.  

\bibliography{aaai2026}


\clearpage
\appendix
\section*{Supplementary Material for ``DualFete: Revisiting Teacher-Student Interactions from a Feedback Perspective for Semi-supervised Medical Image Segmentation''}

\section{Implications of the feedback mechanism}
This work is inspired by Meta Pseudo Labels (MPL)~\cite{pham2021meta} in which a bi-level optimization method is introduced to improve pseudo-labels. Here, we first present theoretical implications followed by MPL's derivation and then identify our distinct contributions. In the final, we present visualizing implications of the feedback in terms of medical image segmentation. 

\subsection{Derivation of the feedback mechanism}
This derivation is adapted from MPL, and we would like interested readers to refer to their work~\cite{pham2021meta}. 

Let $\mathcal{L}_{l}$ denote the supervised loss and $\mathcal{L}_{u}$ the unsupervised loss, defined respectively as: 
\begin{align}
\mathcal{L}_{l}(\theta) &:= \mathbb{E}_{(x,y) \sim \mathcal{D}_{l}} \left[ \ell \left( f(x; \theta), y \right) \right], \nonumber \\
\mathcal{L}_{u}(\theta_{T}, \theta_{S}) &:= \mathbb{E}_{x \sim \mathcal{D}{u}} \left[ \ell \left( f(x; \theta_{S}), \hat{y}^{u} \right) \right], \nonumber
\end{align}
where $\mathcal{D}_{l}$ and $\mathcal{D}_{u}$ denote the labeled and unlabeled datasets, respectively. The pseudo-label $\hat{y}^{u}$ is obtained via teacher prediction: $\hat{y}^{u} = \mathop{\arg\max}_{c} f_{c} \left( x; \theta_{T} \right)$. A bi-level optimization framework is then formulated as:
\begin{align}
    \min_{\theta_{T}} \mathcal{L}_{l} \left( \theta_{S}^{PL}(\theta_{T}) \right), \label{eq:meta:outer} \\
    \text{where } \theta_{S}^{PL}(\theta_{T}) := \mathop{\arg\min}_{\theta_{S}} & \mathcal{L}_{u} (\theta_{T}, \theta_{S}). 
\end{align}
This formulation highlights the dependency of the student $\theta_S^{PL}$ on the teacher $\theta_{T}$ through the pseudo-labels. It defines a meta-objective in which the teacher improves its pseudo-labels by considering the student's performance after updating on those labels. 

Following common practice in meta-learning, the inner optimization $\mathop{\arg\min}_{\theta_{S}}$ is approximated with a single gradient descent step:
\begin{equation}
    \theta_{S}^{PL} \approx \theta_{S} - \eta \Delta \theta_{S} := \theta_{S}', 
\end{equation}
where $\Delta \theta_{S} = \nabla_{\theta_{S}} \mathcal{L}_{u}(\theta_{T}, \theta_{S}; \mathcal{D}_{u}')$ is the stochastic gradient of the unsupervised loss with respect to the student, computed over a mini-batch $\mathcal{D}_{u}'$. Substituting this one-step approximation into Eq.~\ref{eq:meta:outer}, we arrive at the following tractable objective: 
\begin{equation}
    \min_{\theta_{T}} \mathcal{L}_{l} \left(  \theta_{S}'  \right)  = \min_{\theta_{T}} \mathcal{L}_{l} \left(  \theta_{S} - \eta \Delta \theta_{S}   \right). \nonumber
\end{equation}
By applying the chain rule, the gradient of this meta-objective with respect to $\theta_T$ can be computed, allowing the teacher to refine its pseudo-labels: 
\begin{equation}
    \nabla_{\theta_{T}} \mathcal{L}_{l} = - \eta \left( \left[ \nabla_{\theta_{S}} \mathcal{L}_{l} \left( \theta_{S}'; \mathcal{D}_{l}'  \right)   \right]^{\mathsf{T}} \cdot \Delta \theta_{S} \right) \cdot \nabla_{\theta_{T}} \log \mathcal{P}_{\hat{y}^u} \;, 
    \label{eq:meta:obj}
\end{equation}
as presented in \citet{pham2021meta}. Here, $\nabla_{\theta_{S}} \mathcal{L}_{l}(\theta_{S}'; \mathcal{D}_{l}')$ is the stochastic gradient of the supervised loss evaluated at $\theta_{S}'$ on a mini-batch $\mathcal{D}_{l}'$, and $\mathcal{P}_{\hat{y}^u} := \mathcal{P}(\hat{y}^u | x; \theta_{T}, \mathcal{D}_{u}')$ represents the probability assigned by the teacher to the predicted pseudo-label $\hat{y}^u$ over $\mathcal{D}_{u}'$. 

To estimate the inner product between the two gradient vectors with respect to $\theta_S$, we adopt the first-order Taylor approximation: 
\begin{equation}
    \mathcal{L}_{l} \left( \theta_{S} \right) - \mathcal{L}_{l} \left( \theta_{S} - \eta \Delta \theta_{S} \right) \approx \eta \left[ \nabla_{\theta_{S}} \mathcal{L}_{l} \left( \theta_{S}' \right)   \right]^{\mathsf{T}} \cdot \Delta \theta_{S}. 
    \label{eq:first:order:approx}
\end{equation} 
We define the left-hand side of the above approximation as the feedback in our framework. Thus, the error-correcting capability of our feedback mechanism stems from the alignment between two gradients: one reflecting the direction induced by pseudo-label-driven student updates, and the other indicating how the updated student would further reduce the supervised loss. This alignment provides an interpretable and principled basis for adjusting the teacher's pseudo-labels to improve teaching efficacy. 

\begin{figure}[tbp]
    \centering
    \includegraphics[scale=0.825]{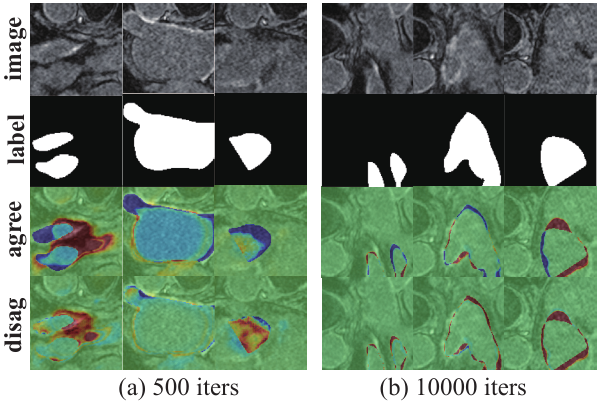}
    \caption{Per-voxel cross-entropy loss changes at 500-th and 10,000 iterations. Label setting: LA 10\%. }
    \label{fig:vis:feed}
\end{figure}
\begin{figure*}[tbp]
    \centering
    \includegraphics[scale=0.85]{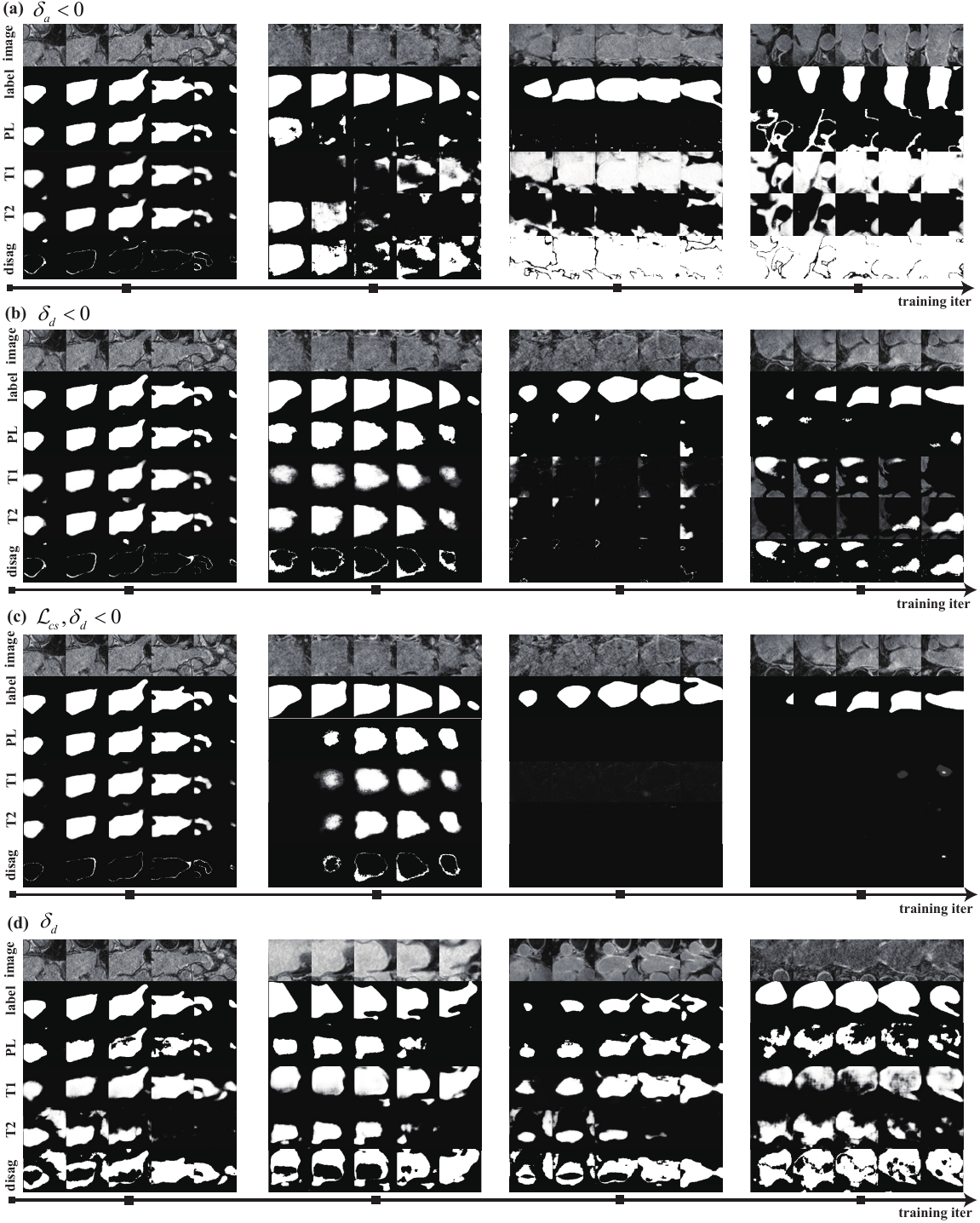}
    \caption{Visualizations for DualFete's qualitative analysis during the second stage. Image, ground-truth label, pseudo-labels (PL), predictions of one teacher (T1), predictions of the other teacher (T2), and disagreement region (disag) are presents. White: foreground, and value=1; black: background, and value=0. }
    \label{fig:vis:quli:evid}
\end{figure*}

\subsection{Distinct contributions of DualFete} 
Eq.~\ref{eq:first:order:approx} presents an approximated solution to MPL in the context of semi-supervised medical image segmentation, corresponding to the single-teacher feedback framework examined in this study. However, this formulation enforces uniform update directions across all voxel probabilities, which is overly rigid and suboptimal for SSMIS. This limitation is further validated by our experimental results, as shown in the second and sixth rows of Table~\ref{tab:fb:ablat}.

To this end, our proposed DualFete framework introduces several methodological innovations:

\textbf{(1) Individualized Student Feedback}: DualFete tailors the student feedback signal for each teacher's predictions, allowing for more adaptive and fine-grained refinement. Thus, our DualFete can circumvent the uniform updating issue while retaining the error-correcting capability of the original design. 

\textbf{(2) Enhanced Cross-teacher Learning}: By adopting a dual-teacher structure, DualFete facilitates collaborative learning between teachers. The cross-teacher learning works synergistically with the dual-teacher feedback, which fosters productive disagreement and prevents error reinforcement. 

\textbf{(3) Human-Aligned Teaching Intuition}: DualFete mirrors human learning manners by modeling two teachers co-instructing a student, receiving individualized feedback, and improving through both self and mutual learning. Despite its ubiquity in reality, to the best of our knowledge, this approach has not been previously explored in the SSMIS literature.

\subsection{Visualizations of feedback}
Figure~\ref{fig:vis:feed} presents feedback visualizations at the 500-th and 10,000-th training iterations, derived from per-voxel cross-entropy loss changes. These heatmaps distinctly capture update patterns induced by pseudo-labels in both agreement (i.e., $\bar{y}^{a}$) and disagreement (i.e., $\bar{y}^{d}$) regions. Two key observations emerge: (1) The divergent performance changes caused by different feedback attributors demonstrate the student's capacity for individualized feedback processing, and (2) The evolving visualization patterns reveal a characteristic learning progression -- early-stage attention to texture features gradually shifts to boundary feature refinement as training progresses. 

\section{Visualizations for DualFete qualitative analysis}
This section supplements the qualitative analysis of DualFete, providing visualizations for intermediate predictions under different constraints, demonstrating in Figure~\ref{fig:vis:quli:evid}. 

When $\delta_{a} < 0$, the two teachers' predictions exhibit increasing complementarity across the entire image, approaching near-complete mutual conflict [Figure~\ref{fig:vis:quli:evid}(a)]. For $\delta_{d} < 0$, the predictions show a progressive decline in boundary confidence, with alternating prediction erosion during training that ultimately collapses to background-only outputs [Figure~\ref{fig:vis:quli:evid}(b) and (c)]. However, when introducing a $\delta_{d} > 0$ constraint alongside $\delta_{d} < 0$, the teachers maintain significantly divergent behaviors while generating highly confident yet distinct predictions [Figure~\ref{fig:vis:quli:evid}(d)]. 

\end{document}